\pdfoutput=1

\documentclass[11pt]{article}

\usepackage[]{acl}

\usepackage{times}
\usepackage{latexsym}
\usepackage{CJKutf8}

\usepackage[export]{adjustbox}

\usepackage{graphicx, subcaption}
\usepackage{multirow}

\usepackage[T1]{fontenc}

\usepackage[utf8]{inputenc}

\usepackage{microtype}

\usepackage{inconsolata}

\title{Prosody in Cascade and Direct Speech-to-Text Translation: \\a case study on Korean \textit{Wh}-Phrases}

\author{First Author \\
  Affiliation / Address line 1 \\
  Affiliation / Address line 2 \\
  Affiliation / Address line 3 \\
  \texttt{email@domain} \\\And
  Second Author \\
  Affiliation / Address line 1 \\
  Affiliation / Address line 2 \\
  Affiliation / Address line 3 \\
  \texttt{email@domain} \\}

\author{Giulio Zhou \quad\quad\quad  Tsz Kin Lam \quad\quad\quad Alexandra Birch \quad\quad\quad Barry Haddow
  \\School of Informatics, University of Edinburgh, United Kingdom\\ \texttt{\normalsize\{giulio.zhou, tlam, a.birch, bhaddow\}@ed.ac.uk } \\
 }
\begin{document}
\maketitle
\begin{abstract}

Speech-to-Text Translation (S2TT) has typically been addressed with \textit{cascade} systems, where speech recognition systems generate a transcription that is subsequently passed to a translation model. While there has been a growing interest in developing \textit{direct} speech translation systems to avoid propagating errors and losing non-verbal content, prior work in direct S2TT has struggled to conclusively establish the advantages of integrating the acoustic signal directly into the translation process.
This work proposes using contrastive evaluation to quantitatively measure the ability of direct S2TT systems to disambiguate utterances where prosody plays a crucial role. Specifically, we evaluated Korean-English translation systems on a test set containing \textit{wh-}phrases, for which prosodic features are necessary to produce translations with the correct intent, whether it's a statement, a yes/no question, a \textit{wh-}question, and more. Our results clearly demonstrate the value of direct translation systems over cascade translation models, with a notable 12.9\% improvement in overall accuracy in ambiguous cases, along with up to a 15.6\% increase in F1 scores for one of the major intent categories. To the best of our knowledge, this work stands as the first to provide quantitative evidence that direct S2TT models can effectively leverage prosody. 
The code for our evaluation is openly accessible and freely available for review and utilisation\footnote{https://github.com/GiulioZhou/contrastive\_prosody}.
\end{abstract}

\section{Introduction}

Speech-to-Text Translation (S2TT) is the task of automatically generating a text translation in a target language given an input speech signal. Traditionally, S2TT has been achieved by concatenating two systems: one in charge of generating an intermediate transcription of the source speech signal and one of translating the intermediate text into a target language. Although such a pipeline, known as \textit{``cascade''} architecture, remains the dominant technology in Speech-to-Text Translation, it has some shortcomings. Firstly, it is affected by error propagation for which errors in the transcription phase are carried over and amplified in the translation phase. Secondly, some information is lost as non-verbal content (e.g. prosody) is discarded from the text. As a potential solution to these issues, ``\textit{direct}'' systems that can perform translation directly from speech signals without needing intermediate transcriptions have emerged in the last few years. \citet{bentivogli-etal-2021-cascade} claim direct systems have an advantage over the cascade architecture by modelling prosody during the translation process. However, there is no conclusive evidence to support this claim as both types of systems have similar overall performances, and current datasets do not regularly include instances where speech signals are necessary to disambiguate the meaning of an utterance, making quantitative analysis on the effect of prosody in S2TT particularly challenging \cite{sperber-paulik-2020-speech, bentivogli-etal-2021-cascade}.

The aim of this paper is to investigate the potential of direct S2TT to effectively leverage non-lexical information, particularly prosody, and quantify their impact. Since identifying ambiguous utterances that rely on prosody for disambiguation is nontrivial, especially in English where sentence structure typically carries more weight than prosodic cues, we focus on Korean \textit{wh}-phrases where the presence of a prosodic boundary distinguishes \textit{wh}-interrogatives from \textit{wh}-indefinites (e.g., \begin{CJK*}{UTF8}{mj}{어디 갔어요}\end{CJK*} (eodi gasseoyo) $\rightarrow$ where did you go?/did you go somewhere?), as well as other interpretations.

In this paper, we (i) introduce a new contrastive evaluation framework for Korean-English S2TT systems, designed for ranking translations of ambiguous utterances containing \textit{wh}-particles; (ii) quantitatively demonstrate the capacity of direct S2TT systems to effectively model prosodic cues from the input, yielding an overall improvement over cascade models of 12.9\% in accuracy for ambiguous utterances, and up to a 15.6\% increase in F1 scores within one of the major intent types; (iii) highlight the limitations of punctuations in disambiguating certain intent types despite being strong signals in distinguishing questions from statements.



\section{Korean Prosody and Wh-Particles}

Prosody refers to the acoustic features that are exhibited across multiple phonetic segments,
also known as suprasegmental features \cite{lehiste1976suprasegmental}. These suprasegmental features can take shape in a multitude of ways. For example, by stressing a single word in a phrase (phrasal stress), by adding pauses or modifying the length of syllables (boundary cues) or by varying the tonal and stress patterns in the utterance (metre) \cite{gerken1998overview}. In an intonational language like Korean, the intended meaning of an utterance is 
often conveyed via intonation and rhythm instead of lexical pitch accents or tones \cite{jun2005prosody, jeon2015prosody}. While prosodic structures in Korean utterances are still debated, there are at least two levels of prosody above the word: the Accentual Phrase (AP) and the Intonation Phrase (IP). The AP is the basic unit for prosodic analysis marked by a tonal pattern THLH which consists of variations of the pitch between low (L) and high (H), with T being either L or H depending on the phrase's initial segment, while the IP consists of one or more APs and a boundary tone on the right edge of the phrase.

\begin{figure}[t]
  \centering

  \begin{subfigure}{\linewidth}
    \includegraphics[width=0.97\linewidth, left]{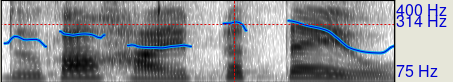}
    \caption{I heard somebody is joining in. (statement)}
    \label{fig:e1}
  \end{subfigure}  
  
  \begin{subfigure}{\linewidth}
    \centering
    \includegraphics[width=\linewidth]{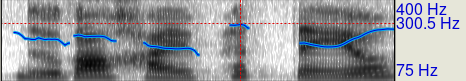}
        \caption{Has somebody joined in? (yes/no question)}
    \label{fig:e2}
  \end{subfigure}

  \begin{subfigure}{\linewidth}
    \centering
    \includegraphics[width=\linewidth]{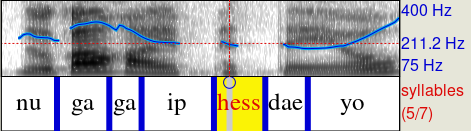}
        \caption{Who joined in? (\textit{wh-}question)}
    \label{fig:e3}
  \end{subfigure}  
  
  \caption{
  An example in the ProSem dataset: Based on the intent, the transcription ``\begin{CJK*}{UTF8}{mj}{누가 가입했대요}\end{CJK*}'' (nuga gaiphessdaeyo) can be mapped to a different pair of recording and translation, see (a), (b), and (c). The ``blue” lines on the spectrogram, i.e., the recording, are the pitch (F0) contours.}
  \label{fig:example1}
\end{figure}  

Korean \textit{wh-}particles are an example of a linguistic phenomenon where the tonal patterns and IP boundary tones are necessary to disambiguate the meaning of the utterance, as otherwise they can be interpreted as both interrogative particles or indefinite pronouns (e.g. \begin{CJK*}{UTF8}{mj}{누구}\end{CJK*}'' (nugu) $\rightarrow$ ``who'' / ``somebody''). Figure \ref{fig:example1} shows the pitch contours for the recordings of the utterance ``\begin{CJK*}{UTF8}{mj}{누가 가입했대요}\end{CJK*}'' (nuga gaiphessdaeyo). By varying the boundary tone H+L\%, H+LH\%, and L+H\%, the utterance can be interpreted as a statement, yes/no question or \textit{wh-}question respectively.

\section{Contrastive Evaluation}\label{sec:cont}

Contrastive evaluation is an automatic accuracy-based evaluation technique that measures the capability of a system to distinguish correct from incorrect outputs.
This is achieved by asking a generative model $\theta$ to score and rank a set of predefined outputs, each containing a correct and a contrastive utterance (e.g., \textit{``the cat sleeps''} vs. \textit{``the cat sleep''} \cite{linzen2016assessing}). Following previous work \cite{sennrich-2017-grammatical, vamvas-sennrich-2021-limits}, we define the score of an utterance as the sum of the target token log probabilities normalised by the length of the full target sequence $Y$:

$\textrm{score}(Y|X,\theta)=\frac{1}{|Y|}\sum\limits_{i=1}^{|Y|}\textrm{log}p_\theta(y_i|X,y_{<i})$

where $X$ is the input signal, $|Y|$ the target sequence length and $\theta$ the evaluated model.  

In this work, we perform contrastive evaluation of cascade and direct S2TT systems on Korean \textit{wh}-phrases. Since multiple prosodic realisations can occur per utterance (as in Figure \ref{fig:example1}), in contrast to previous work where only one contrastive utterance per example was available, we consider a model having correctly identified the intended translation only if its score is higher compared to the score of all the possible incorrect translations.  
In addition to the general accuracy of the model in identifying the correct translation, we report contrastive precision, recall and F1 scores of the systems on the various \textit{wh-}phrases' intent types.

\section{Experimental Setting}\label{sec:set}

\begin{table}[t]
\begin{tabular}{lcllc}
\cline{1-2} \cline{4-5}
\multicolumn{1}{c}{Intent} & \#   &  & \multicolumn{1}{c}{Wh-particle} & \#    \\ \cline{1-2} \cline{4-5} 
Statement                  & 1085 &  & Who                             & 1,895 \\
Yes/no Q                   & 1047 &  & What                            & 877   \\
Wh-Q                       & 849  &  & Where                           & 199   \\
Rhetorical Q               & 302  &  & When                            & 172   \\
Commands                   & 175  &  & How                             & 163   \\
Requests                   & 56   &  & How many                        & 246   \\ \cline{4-5} 
Rhetorical C               & 38   &  &                                 &       \\ \cline{1-2}
\end{tabular}
\caption{Number of utterances in Prosem per \textit{wh-}particle and intent type.}
\label{tab:prosem}
\end{table}

In our experiment, we adopted the ProSem corpus \cite{cho2019prosody} as the contrastive evaluation test set. Originally designed for Spoken Language Understanding, this corpus consists of 3552 utterances recorded by two Korean native speakers of a different gender. All the utterances make use of one of the six Korean \textit{wh-}particles
and are further classified into seven intent categories: statements, yes/no questions, \textit{wh-}questions, rhetorical questions, commands, requests, and rhetorical commands, with the first three categorised as major intent types. Table \ref{tab:prosem} shows the number of utterances per intent type and \textit{wh-}particle in the Prosem dataset. In the dataset, there are a total of 1292 distinct transcriptions, each associated with up to 4 utterances of a different intent. 
Each recorded utterance in the dataset is thus paired to a gold translation, as well as a number of incorrect ones that are associated with recordings of the same transcription (but with different prosody). For example, in the recording in Figure \ref{fig:e1} the correct translation is \textit{``I heard somebody is joining in.''} while the incorrect/contrastive ones are \textit{``Has somebody joined in?''} and \textit{``Who joined in?'' }.

For our experiments, we utilise state-of-the-art pretrained models. Specifically, we use Open AI's Whisper models \cite{radford2022robust} for both the S2TT direct systems and the ASR components in the cascade systems, reporting results obtained from all the provided multilingual models. As for the MT component in the cascade systems, we make use of the Korean-English baseline model provided for the Tatoeba challenge \cite{tiedemann-2020-tatoeba}, trained on approximately 34.5M Opus MT parallel data \cite{tiedemann-thottingal-2020-opus}. 


\section{Results}\label{sec:res}

\begin{figure}[t]
\centering
  \includegraphics[width=\linewidth]{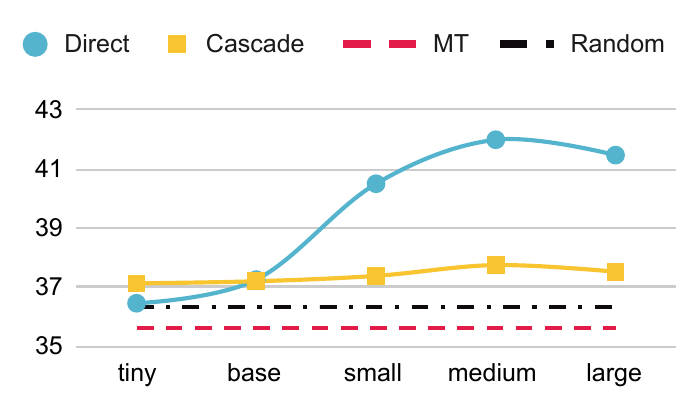}
\caption{Contrastive evaluation accuracy $\uparrow$ scores on ProSem for direct (blue) and cascade (yellow) S2TT systems by varying the size of the Whisper model, along with Random selection (black) and an MT system that has gold transcriptions as input (red).}
  \label{fig:whisp}
\end{figure}  

\subsection{Contrastive Evaluation Accuracy} 
Figure \ref{fig:whisp} shows the results of the contrastive evaluation, along with the average accuracy of randomly selecting one of the 2-4 potential translations.
As expected, the performance of both cascade and direct systems exhibits an upward trend with increasing model size. Notably, the direct systems outperformed both the MT and cascade systems, with the ``medium'' direct system exhibiting an improvement of 6.4\% and 4.3\% in accuracy respectively.

In contrast, the MT model with gold transcription as input failed to surpass random selection in performance due to its inability to distinguish between different translations effectively when presented with the same transcription. On the other hand, despite relying on the aforementioned MT model, the cascade systems managed to achieve scores surpassing random selection, with an improvement of up to 2.1\% observed in the Whisper ``medium'' system. This improvement can be attributed to the inclusion of punctuation marks in the transcriptions, which are absent in the gold transcriptions, that aid in disambiguating questions from statements.

\begin{table*}[t]
\centering
\begin{tabular}{cccccccc}
\hline
   & \multirow{2}{*}{\begin{tabular}[c]{@{}c@{}}Direct\\ medium\end{tabular}} & \multicolumn{2}{c}{Cascade medium}                          & \multicolumn{2}{c}{MT}            & \multirow{2}{*}{Random} & \multirow{2}{*}{\begin{tabular}[c]{@{}c@{}}Wh-Q\\ Random\end{tabular}} \\ \cline{3-6}
   &                         & \multicolumn{1}{c}{W/O} & \multicolumn{1}{c}{W} & \multicolumn{1}{c}{W/O} & W  &                         &                                                                        \\ \hline
Ambiguous  & 48.9                   & 36.4                    & 39.2                    & 36.5                     & 39.3 & 32.3                   & 42.8                                                                  \\
Unambiguous & 33.6                  & 34.7                     & 36.0                    & 34.6                     & 40.8  & 41.3                   & 28.6                                                                  \\ \hline
\end{tabular}
\caption{Contrastive evaluation accuracy $\uparrow$ scores on ambiguous and unambiguous contrastive sets for systems without (W/O) and with (W) question marks in the input, and pure and \textit{wh-}question biased random selection. }
  \label{tab:ambiguousquestion}
\end{table*}

\subsection{Effect of Punctuation}
To better understand the disparity in performance between direct and cascade systems, we conducted an analysis to assess the role of punctuation within the MT inputs. To do so, we added question marks to the ProSem gold transcriptions based on the intents of the correct translations. Subsequently, we categorised the contrastive sets into two distinct groups: ``Ambiguous'' and ``Unambiguous'', where the latter are the ones where punctuation alone is sufficient to discern the correct intention among the options considered. Figure \ref{fig:example1} illustrates examples for both ambiguous and unambiguous contrastive sets. The contrastive set where ``statement'' (Figure \ref{fig:e1}) is the correct translation is an example of an unambiguous set because the absence of a question mark in the transcription ``\begin{CJK*}{UTF8}{mj}{누가 가입했대요}\end{CJK*}'' is sufficient to identify the correct intent as a ``statement'' as both yes/no and \textit{wh-}questions contain question marks. On the other hand, in the scenario where the correct translation corresponds to the utterance with \textit{wh}-question intent type (Figure \ref{fig:e3}), the set becomes ambiguous, as ambiguity arises because both yes/no and \textit{wh-}questions share the same transcription ``\begin{CJK*}{UTF8}{mj}{누가 가입했대요?}\end{CJK*}''. In total, we identified 1602 unambiguous and 1950 ambiguous sets.

\begin{figure}%
    \centering
   \includegraphics[width=6cm]{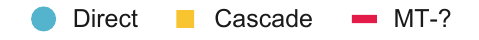} %
   
    \subfloat[Ambiguous]{{\includegraphics[width=3.7cm]{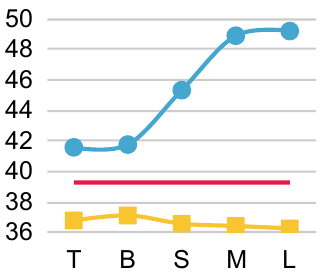}}\label{fig:ambiguous}}
     \subfloat[Unambiguous]{{\includegraphics[width=3.8cm]{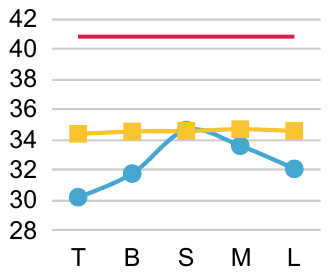} }\label{fig:unambiguous}}

\caption{Contrastive evaluation accuracy $\uparrow$ for direct (blue), cascade (yellow) S2TT systems with different Whisper model sizes, and MT with gold transcriptions augmented with question marks (MT-?, red) on ambiguous and unambiguous contrastive sets.}
  \label{fig:avsuna}
\end{figure}

\subsubsection{Accuracy on Ambiguous Examples}
First, we performed the contrastive evaluation on the previously illustrated ambiguous and unambiguous sets. Figure \ref{fig:ambiguous} shows that, on ambiguous contrastive sets, all direct systems consistently outperform their cascade counterparts and even surpass the MT system, which has access to gold transcriptions. The gap between the direct and cascade systems is notably wider compared to the overall performance shown in Figure \ref{fig:whisp}, with differences reaching up to 12.9\% for the ``large'' model, supporting the hypothesis that direct models are capable of modelling acoustic signals to handle ambiguous utterances effectively. On the other hand, Figure \ref{fig:unambiguous} shows that the augmented gold MT model, which serves as an upper bound for the cascade systems, outperforms the best-performing direct model by 6.2\% in accuracy, illustrating that punctuation is an effective convoy for certain prosodic information. The effectiveness of punctuation is reflected in the performance of cascade systems themselves, which, except for the ``small'' model, outperform the direct systems. It's worth noting that all systems, despite their strengths, did not achieve the anticipated levels of performance on the unambiguous contrastive sets. This can be attributed to the ambiguity caused by the absence of mandatory question marks in modern Korean. The resulting inconsistencies in question mark usage within existing training data, where questions may lack proper punctuation, contribute to errors in the models' understanding of sentence types.

\subsubsection{Adding/Removing Punctuation}
To further explore the effect of punctuation, we manipulated MT inputs by either removing question marks from the ASR transcription or augmenting the gold transcription. In Table \ref{tab:ambiguousquestion}, we present results for systems with and without question marks, including accuracy for pure random selection (``Random'') and an additional random baseline biased towards selecting \textit{wh-}question intent types (i.e., choosing a \textit{wh-}question if it's an option, and selecting randomly otherwise) to simulate better the behaviour of the systems (``Wh-Q Random'', see Sec \ref{sec:intent}). Despite MT-based systems outperforming pure random selection, they fall short of surpassing the ``Wh-Q Random'' baseline on ambiguous sets as the input transcription lacks sufficient information to disambiguate the correct intent. 

For unambiguous examples, introducing question marks in the MT input results in a significant improvement in scores. Notably, the MT system with gold transcription outperforms the direct S2TT model in handling these examples. However, none of the systems seem to perform better than random selection, a limitation attributed to a bias towards \textit{wh-}questions. Overall, these findings align with our previous results, emphasising the advantage of direct S2TT models over text-based systems due to their ability to leverage prosodic information for disambiguating sentences. While punctuation aids in differentiating questions from statements, it remains insufficient to resolve all instances of ambiguity.


  


  

\begin{figure}[t]
\centering
\includegraphics[width=6cm]{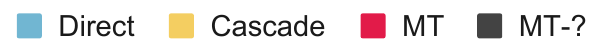}
\includegraphics[width=.55\columnwidth]{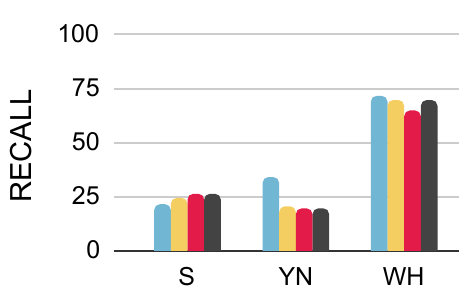}
\includegraphics[width=.43\columnwidth]{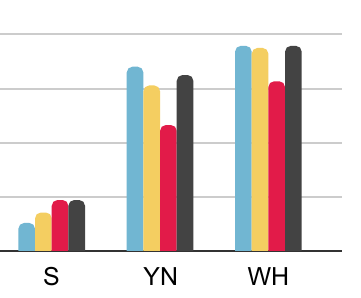}

\includegraphics[width=.55\columnwidth]{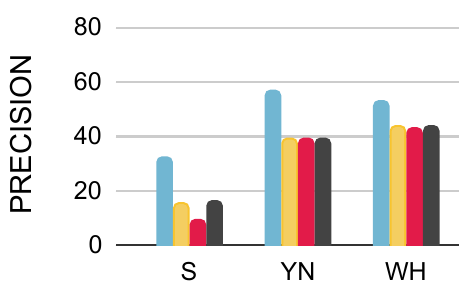}
\includegraphics[width=.43\columnwidth]{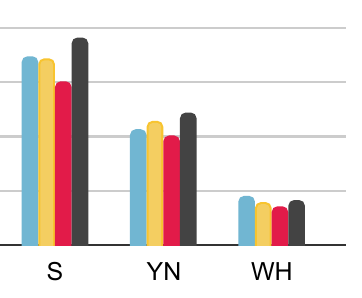}

\includegraphics[width=.55\columnwidth]{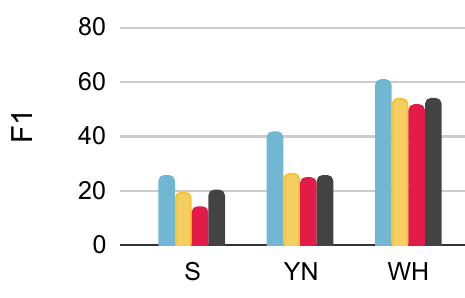}
\includegraphics[width=.42\columnwidth]{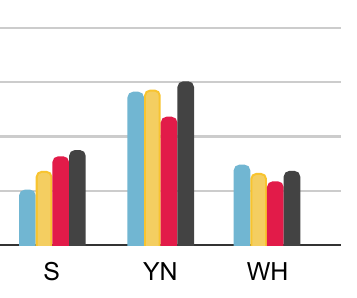}

\hspace{1.2cm}{\small(a) Ambiguous} \hspace{0.9cm}{\small (b) Unambiguous}
\caption{Contrastive evaluation recall, precision and F1 $\uparrow$ scores on ambiguous and unambiguous contrastive pairs for each intent major type: statements (S), yes/no questions (YN), \textit{wh-}questions (WH). Direct and cascade systems based on Whisper ``medium'', and MT systems with and without gold question marks.}
\label{fig:unseenresults}\label{fig:intent}
\end{figure}





\subsection{Intent Disambiguation}\label{sec:intent}
While Figure \ref{fig:whisp} and \ref{fig:ambiguous} demonstrate the advantages of preserving acoustic signals during the translation process, it's important to note that the overall accuracy achieved by all systems remains relatively low. 
Figure \ref{fig:intent} reveals a significant challenge common to all systems when it comes to disambiguating statements, as they achieve a recall score of less than 25\% in this category. In contrast, the highest recall scores are consistently observed in the \textit{wh-}questions intent category. The low recall score for yes/no questions and the subpar precision for \textit{wh-}questions, two intent types that are indistinguishable for MT-based systems, indicate a distinct bias towards the \textit{wh-}question type. This bias can be attributed to the primary use of \textit{wh-}particles in the Korean language for forming \textit{wh-}questions.

Overall, on ambiguous contrastive sets, the direct model outperforms the other two systems in terms of F1 scores across all major intent categories, achieving improvements of up to 15.5\% in the case of yes/no questions. However, on unambiguous sets, the direct model's performance is comparable to cascade models in question categories but falls short on statements, where its recall is notably low. This performance gap on statements may be due to the inherent challenge of accurately capturing the nuanced prosody and context associated with statements, which direct models may struggle to discern effectively. Full results and confusion matrices are reported in Appendix \ref{app:full}.

\section{Conclusion}
The objective of this paper was to test whether direct S2TT systems could take advantage of the prosodic information contained in the speech signal. To achieve this, we conducted quantitative analyses focused on Korean \textit{wh-}particles which can represent either \textit{wh-}interrogatives or \textit{wh-}indefinites encompassing a range of intents in accordance with the input acoustic features. 
Our contrastive evaluation results provide compelling evidence that the direct S2TT systems outperform the cascade systems in overall accuracy and F1 score across all the major intent types on ambiguous utterances. Cascade systems perform better than random primarily thanks to the inclusion of punctuation in the transcriptions. However, it's essential to note that while punctuation marks play a valuable role in aiding disambiguation, they are not sufficient to resolve all types of intents, emphasizing the importance of considering prosody in S2TT systems.

\section*{Limitations}

While our study has yielded positive results, it is essential to acknowledge several limitations. Firstly, the contrastive evaluation approach in this study diverges from previous work in that it was not conducted with minimally different utterances. The set of possible translations used here differs significantly in structure and, to some extent, vocabulary. This variation may potentially influence the resulting scores, despite being normalised.
Secondly, the findings of this research may not be readily generalisable beyond the specific context of Korean \textit{wh-}particles. To examine different linguistic phenomena in various language pairs, specific contrastive datasets will need to be meticulously crafted. As previously discussed, this process poses a significant challenge.
Lastly, despite employing state-of-the-art models, the overall accuracy observed in the contrastive evaluation remains relatively low. This suggests that there is substantial room for improvement within speech translation systems, reflecting the ongoing development needs in this field.


\section*{Acknowledgements}
This work was supported in part by the Centre for Doctoral Training in Natural Language Processing, funded by the UKRI (grant EP/S022481/1); and also by the UKRI under the UK government’s Horizon Europe funding guarantee (10039436 – UTTER) and by the University of Edinburgh, School of Informatics.

    \bibliography{anthology,custom}

\appendix

\section{Korean \textit{wh-}particles}\label{app:data}

\begin{table}[t]
\centering

\resizebox{\columnwidth}{!}{
\begin{tabular}{lccc}
\hline
\multicolumn{2}{c}{Wh-Particle} & Interrogative &Indefinite   \\ \hline
      \begin{CJK*}{UTF8}{mj}{뭐}\end{CJK*}     & mwo                & what             & something       \\
    \begin{CJK*}{UTF8}{mj}{누구} \end{CJK*}      & nugu               & who                & someone         \\
        \begin{CJK*}{UTF8}{mj}{언제}  \end{CJK*} & eonje              & when              & some time       \\
       \begin{CJK*}{UTF8}{mj}{어디} \end{CJK*}   & eodi               & where              & some place      \\
        \begin{CJK*}{UTF8}{mj}{어떻게} \end{CJK*}  & eotteohge          & how                 & somehow         \\
        \begin{CJK*}{UTF8}{mj}{몇}\end{CJK*}   & myeot                & how many    &       some  \\ \hline
\end{tabular}}
\caption{Korean \textit{wh-}Particles and English \textit{wh-}interrogatives/indefinite pronouns in the ProSem dataset.}\label{tab:wh}
\end{table}


Table \ref{tab:wh} shows Korean \textit{wh-}particles and their English translations. The particle \begin{CJK*}{UTF8}{mj}{왜}\end{CJK*} (wae, why) is not present in the ProSem dataset as it is rarely used as a quantifier. On the other hand, \begin{CJK*}{UTF8}{mj}{몇}\end{CJK*} (myeot, how many) is used instead despite not being technically a \textit{wh-}particle. 

\section{General Performance}\label{app:general performance}
We present the SacreBLEU\footnote{nrefs:var|case:mixed|tok:13a|smooth:exp|version:1.5.1} \cite{post-2018-call} score and the Character Error Rate \cite[CER]{morris2004} of the systems to assess their general performance in the translation and transcription tasks respectively. In addition to the results on the ProSem testset, we provide general performance on the kosp2e \cite{cho21b_interspeech} test set. As shown in Table \ref{tab:bleu}, the results align with expectations, demonstrating that Whisper's performance improves with model size for both translation and recognition tasks on both test sets. The direct systems perform well on both test sets with BLEU scores up to 21.1 and 21.4 on the kosp2e and ProSem test sets respectively. As for the cascade systems, it is worth noting that the MT on gold transcription serves as an upper benchmark for the performance of the cascade systems. However, we can see that all the cascade systems achieve a higher BLEU score on ProSem compared to the base MT model. As discussed in Section \ref{sec:res}, this is mainly due to the lack of punctuation in the transcription. By augmenting the model with question marks, we can see a drastic increase in BLEU score reaching 15.0, outperforming the cascade systems. Moreover, by comparing the CER scores on the two test sets, we observe that they are generally higher on the ProSem test set. This suggests that the utterances in the ProSem test set may be considered out-of-domain compared to more general test sets, contributing to the higher CER scores.
\begin{table}[t]
\centering
{\small 
\begin{tabular}{lcll}
\hline
Model                    & Size & kosp2e & ProSem \\ \hline
\multirow{5}{*}{Direct}  & T    & 1.0   & 5.3   \\
                         & B    & 4.7   & 10.2  \\
                         & S    & 13.0     & 17.4  \\
                         & M    & 19.4 & \textbf{21.4}  \\
                         & L    & \textbf{21.1}  & 19.6  \\ \hline
\multirow{5}{*}{Cascade} & T    & 10.6 (16.2)  & 10.9 (27.0) \\
                         & B    & 12.3 (12.1)  & 12.2 (22.3) \\
                         & S    & 13.9 (9.1)  & 13.3 (16.3) \\
                         & M    & 14.9 (7.3) & 14.1 (\textbf{13.9}) \\
                         & L    & 15.2 (\textbf{6.6}) & 14.3 (\textbf{13.9}) \\ \hline
MT                       &      & 14.2  & 7.2 / 15.0   \\ \hline
\end{tabular}
}
\caption{BLEU $\uparrow$ scores for Whisper-S2TT (Direct), Whisper-ASR+MT (Cascade) and MT with gold transcriptions on the kosp2e and ProSem (without and with additional punctuation) test sets. Model sizes: tiny (T), base (B), small (S), medium (M) and large (L). CER $\downarrow$ for Whisper-ASR in brackets.} 
\label{tab:bleu}
\end{table}

\section{Full Intent Disambiguation Results}\label{app:full}

\begin{figure}[t]
  \centering

  \begin{subfigure}{\linewidth}
    \centering
    \includegraphics[width=\linewidth]{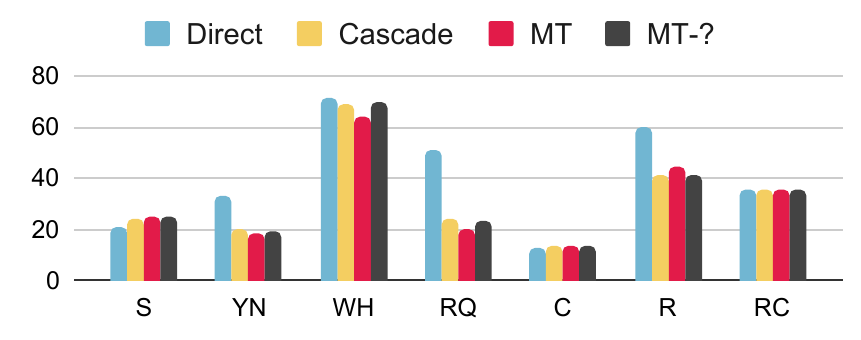}
    \caption{Recall}
    \label{fig:recallfulla}
  \end{subfigure}  
  
  \begin{subfigure}{\linewidth}
    \includegraphics[width=\linewidth,right]{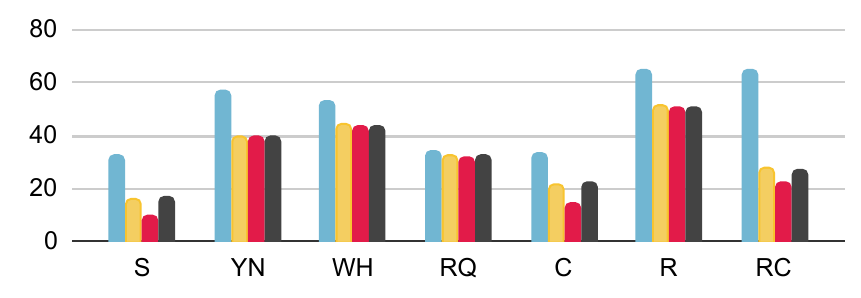}
    \caption{Precision}
    \label{fig:precisionfulla}
  \end{subfigure}

  \begin{subfigure}{\linewidth}

    \includegraphics[width=\linewidth,right]{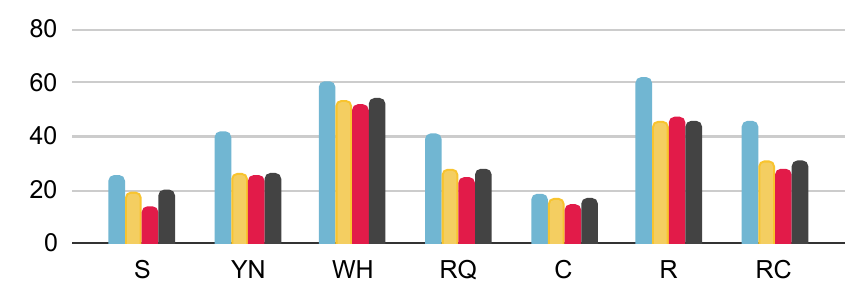}
    \caption{F1}
    \label{fig:f1fulla}
  \end{subfigure}  
  
  \caption{Contrastive evaluation recall, precision and F1 $\uparrow$ scores con \textit{ambiguous} sets for direct and cascade Whisper ``medium'', and Machine Translation systems, for each intent type.}  
  \label{fig:intentfulla}
\end{figure}

\begin{figure}[t]
  \centering

  \begin{subfigure}{\linewidth}
    \centering
    \includegraphics[width=\linewidth]{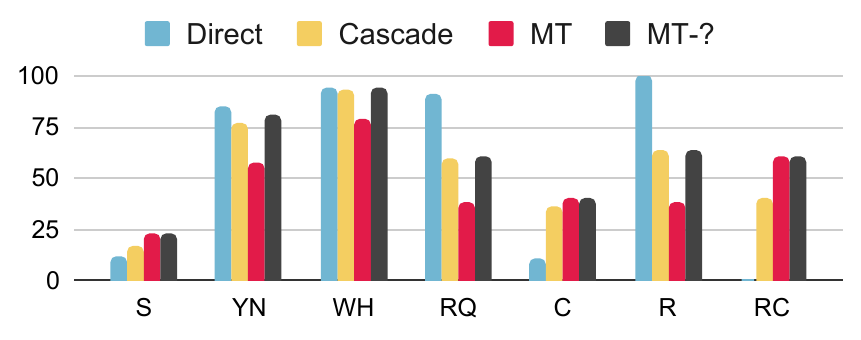}
    \caption{Recall}
    \label{fig:recallfullua}
  \end{subfigure}  
  
  \begin{subfigure}{\linewidth}
    \includegraphics[width=\linewidth,right]{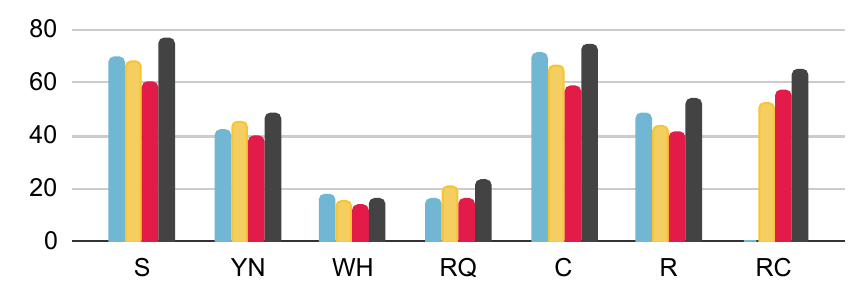}
    \caption{Precision}
    \label{fig:precisionfullua}
  \end{subfigure}

  \begin{subfigure}{\linewidth}

    \includegraphics[width=\linewidth,right]{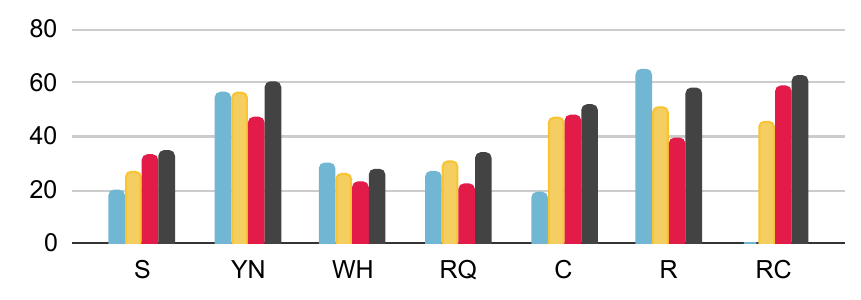}
    \caption{F1}
    \label{fig:f1fullua}
  \end{subfigure}  
  
  \caption{Contrastive evaluation recall, precision and F1 $\uparrow$ scores con \textit{unambiguous} sets for direct and cascade Whisper ``medium'', and Machine Translation systems, for each intent type.}  
  \label{fig:intentfullua}
\end{figure}

\begin{figure}[t]
  \centering

  \begin{subfigure}{\linewidth}
    \centering
    \includegraphics[width=\linewidth]{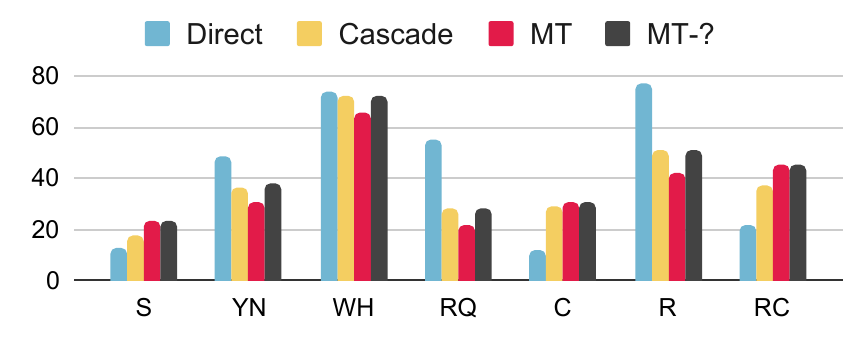}
    \caption{Recall}
    \label{fig:recallfull}
  \end{subfigure}  
  
  \begin{subfigure}{\linewidth}
    \includegraphics[width=\linewidth,right]{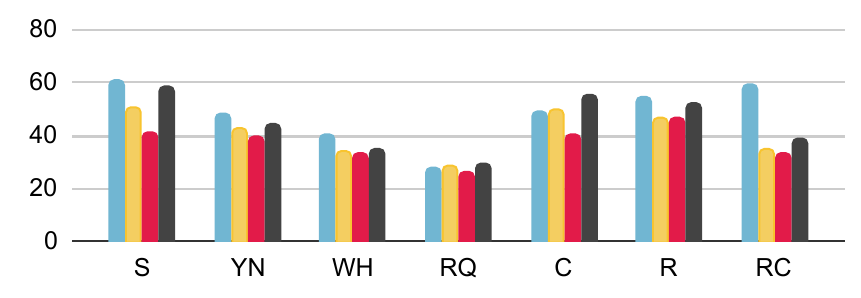}
    \caption{Precision}
    \label{fig:precisionfull}
  \end{subfigure}

  \begin{subfigure}{\linewidth}

    \includegraphics[width=\linewidth,right]{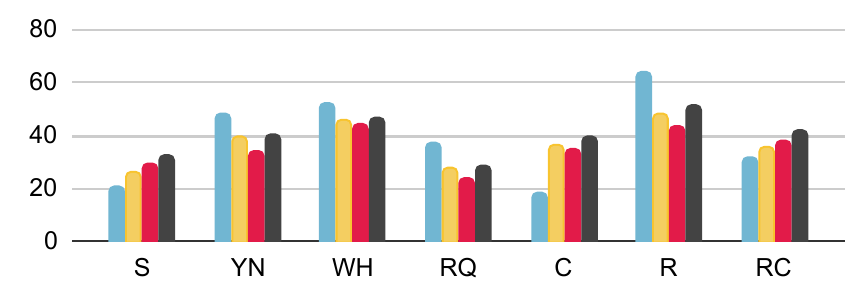}
    \caption{F1}
    \label{fig:f1full}
  \end{subfigure}  
  
  \caption{Overall contrastive evaluation recall, precision and F1 $\uparrow$ scores on ProSem for direct and cascade Whisper ``medium'', and Machine Translation systems, for each intent type.}  
  \label{fig:intentfull}
\end{figure}


  


    \begin{figure*}
        \centering
        \begin{subfigure}[b]{0.48\textwidth}
            \centering
            \includegraphics[width=\textwidth]{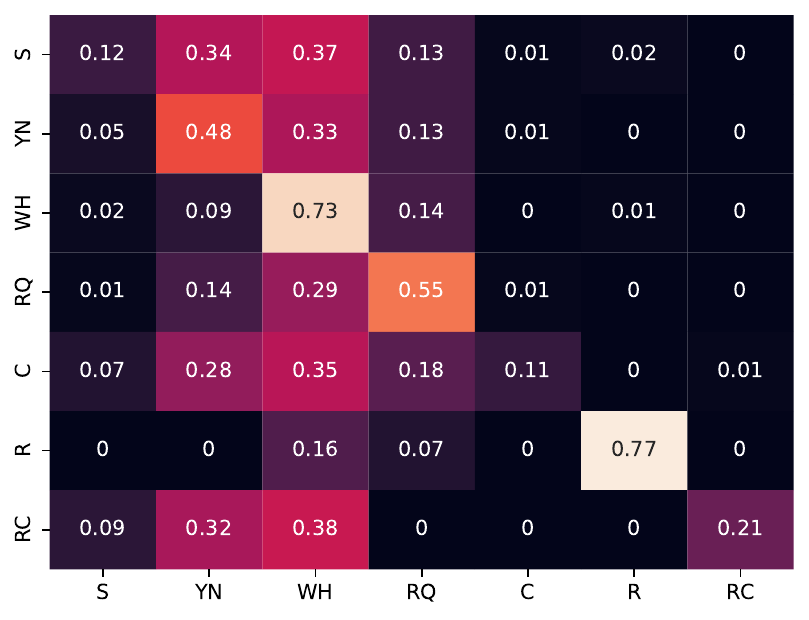}
            \caption{Direct}%
            \label{fig:cm_direct}
        \end{subfigure}
        \hfill
        \begin{subfigure}[b]{0.48\textwidth}  
            \centering 
            \includegraphics[width=\textwidth]{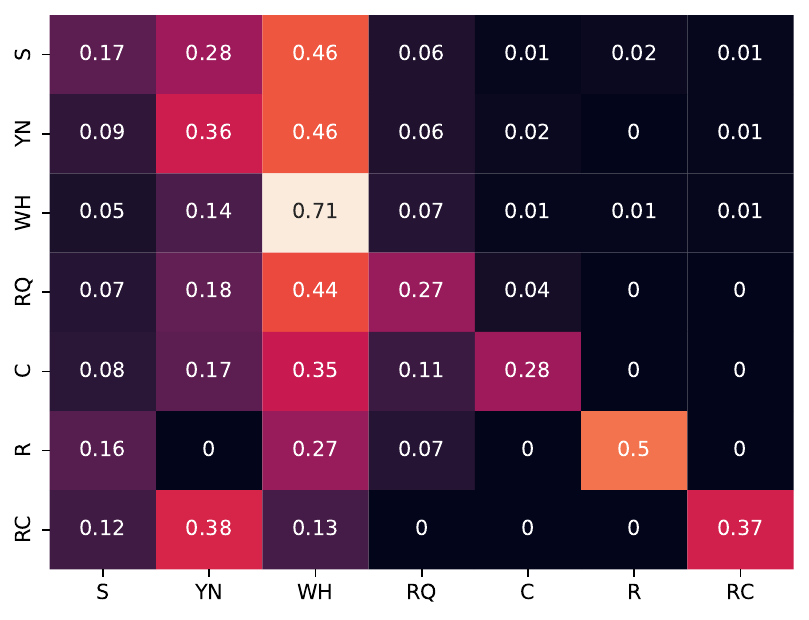}
        \caption{Cascade}%
            \label{fig:cm_cascade}
        \end{subfigure}
        \vskip\baselineskip
        \begin{subfigure}[b]{0.48\textwidth}   
            \centering 
            \includegraphics[width=\textwidth]{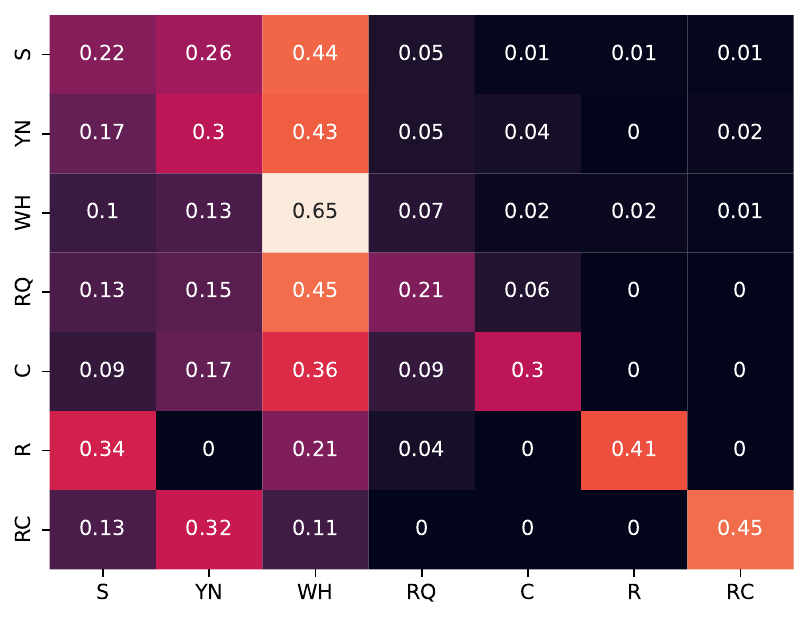}
            \caption{MT}%
            \label{fig:cm_mt}
        \end{subfigure}
        \hfill
        \begin{subfigure}[b]{0.48\textwidth}   
            \centering 
            \includegraphics[width=\textwidth]{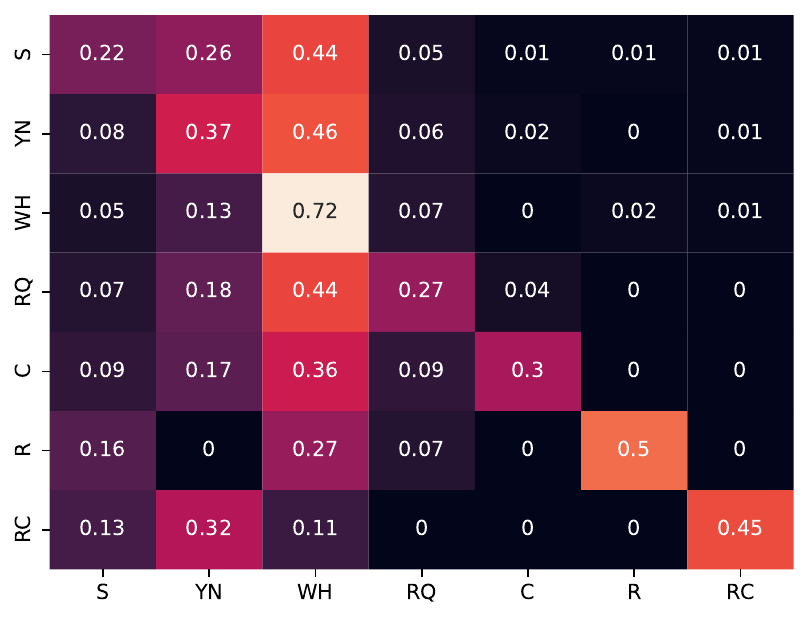}
            \caption{MT-?}%
            \label{fig:cm_mtq}
        \end{subfigure}
        \caption{Normalised confusion matrices for Whisper ``medium'' direct and cascade, and Machine Translation (MT) systems with and without additional punctuation. Classes: statements (S), yes/no questions (YN), \textit{wh-}questions (WH), rhetorical questions (RQ), commands (C), requests (R), and rhetorical commands (RC).}
        \label{fig:confusion}
    \end{figure*}

Figure \ref{fig:intentfulla}, \ref{fig:intentfullua} and \ref{fig:intentfull} shows the recall, precision and f1 scores for the models on all the intent types (statements (S), yes/no questions (YN), wh-questions (WH), rhetorical questions (RQ), commands (C), requests (R), and rhetorical commands (RC)). In the context of ambiguous contrastive sets (Figure \ref{fig:intentfulla}), the direct system consistently outperforms other models across all intent types, showcasing superior performance across all metrics. On unambiguous sets, the direct systems excel primarily in achieving high recall scores for questions (yes/no questions, \textit{wh-}questions, rhetorical commands, and requests). However, for non-question intent types, the direct systems exhibit recall scores often below 12\%, plummeting as low as 0\% for rhetorical commands. This differentiation is reflected in the overall results (Figure \ref{fig:intentfull}), where the direct system surpasses text-based models in terms of F1 scores specifically for questions.

Figure \ref{fig:confusion} offers a closer look at the confusion matrices for the systems during the intent disambiguation task in contrastive evaluation. As detailed in Section \ref{sec:res}, it's evident that all models display a notable bias toward the \textit{wh-}question intent type, a tendency that is particularly pronounced in cascade and MT systems. Notably, the MT model, when not augmented with additional punctuation, exhibits a stronger inclination toward interpreting utterances as statements, especially evident in requests, where the incorrect selection of statements significantly decreases when punctuation is added (from 34\% to 16\%).
Overall, the confusion matrices shed light on the challenges faced by text-based systems in effectively disambiguating intent, indicating a preference for interpreting utterances as one of the three major intent types.

\section{Vanilla Models} \label{app:scratch}

\begin{table}[t]
\centering
\begin{tabular}{llrr}
\hline
\multicolumn{1}{c}{Dataset} & \multicolumn{1}{c}{Split} & \#     & hs  \\ \hline
ProSem                      & test                          & 7104   & 7   \\ \hline
\multirow{3}{*}{kosp2e}     & train                     & 106653 & 257 \\
                            & dev                       & 1266   & 2   \\
                            & test                      & 2320   & 4   \\ \hline
ClovaCall                   & train                     & 59662  & 50  \\ \hline
Korean                      & train                     & 125226 &     \\
Parallel Corpora                & dev                       & 1720   &     \\ \hline \hline
\multirow{2}{*}{S2TT}         & train                     & 106652 & 257 \\
                            & dev                       & 1266   & 2   \\ \hline
\multirow{2}{*}{ASR}        & train                     & 166315 & 307 \\
                            & dev                       & 1266   & 2   \\ \hline
\multirow{2}{*}{MT}         & train                     & 231879 &     \\
                            & dev                       & 2986    &     \\ \hline
\end{tabular}
\caption{Datasets sizes in number of utterances/parallel sentences and recordings time in hours. Bottom half shows the data sizes used for training the direct S2TT, ASR and MT systems. }\label{tab:stat}
\end{table}

In this section, we report the results for smaller direct and cascade S2TT systems trained from scratch. To train our models, we used three distinct datasets: kosp2e \cite{cho21b_interspeech}, Korean Parallel corpora \cite{park-etal-2016-korean} and ClovaCall \cite{ha2020clovacall}.
The kosp2e dataset was used to train all the systems as it contains speech signals, transcriptions and translation required to train direct S2TT, ASR and MT models. ClovaCall was used with kosp2e to train ASR systems, while the Korean Parallel corpora were used for MT systems as described in Section \ref{sec:set}. Table \ref{tab:stat} shows the statistics of the datasets used for training the systems. We used \textit{fairseq S2T} \cite{wang-etal-2020-fairseq} implementations for the S2TT and ASR models, with \textit{``s2t transformer''} architectures and default training settings. In addition, we report results for a direct S2TT model with an ASR-initialised encoder. All results are the average of four different seeds.

\subsection{Results}

\begin{table}[t]

\resizebox{\columnwidth}{!}{
\begin{tabular}{lcll}
\hline
\multicolumn{1}{c}{Model}      & Size & \multicolumn{1}{c}{kosp2e} & \multicolumn{1}{c}{ProSem} \\ \hline
\multirow{2}{*}{Direct}            & S    & 2.0                       & 0.7                       \\
                               & M    & 2.1                       & 0.5                      \\
\multirow{2}{*}{Direct+ASR init} & S    & 9.1                       & 1.6                       \\
                               & M    & 8.8                       & 1.6                       \\ \hline
\multirow{2}{*}{Cascade}      & S    & 0.2 (88.9)               & 0.1 (125.6)              \\
                               & M    & 0.2 (88.6)               & 0.1 (127.4)                           \\ \hline   \hline
MT                             &      & \textbf{19.7}                      & 9.5 / \textbf{11.4}   \\ \hline   
\end{tabular}}
\caption{BLEU $\uparrow$ scores and CER $\downarrow$ (in brackets) for direct and cascade Speech-to-Text Translation systems trained from scratch with architecture small (S) and medium (M), and MT models (without/with gold punctuation on the ProSem test set).}
\label{tab:bleu2}
\end{table}

Results in Table \ref{tab:bleu2} show the general performance of the direct and cascade systems trained from scratch. Compared to the results for whisper-based models in Section \ref{sec:res}, the base direct and cascade systems could not provide satisfactory outputs on either test sets. However, despite the poor performance of the ASR models (CER > 88\%), when used to initialise the direct S2TT models, they improved drastically the latter's performance, with an increase of 7.1 and 6.7 points in BLEU for the small and medium models respectively on the kosp2e test set. It's worth noting that the MT system, despite being trained on a notably smaller dataset compared to the OpusMT model, managed to achieve a high BLEU score on the kosp2e test set. This can be attributed to its training on in-domain data, underlining the impact of domain-specific training in enhancing performance.


\begin{table}[t]
\centering
\begin{tabular}{lc}
\hline
\multicolumn{1}{c}{Model} & Accuracy \\ \hline
Random                    & 36.3    \\
MT                        & 35.4    \\
MT-?                        & 39.4    \\
Cascade                   & 36.4    \\
Direct                    & 36.1    \\
Direct+ASR init           & \textbf{39.8}    \\ \hline

\end{tabular}
\caption{Contrastive evaluation accuracy $\uparrow$ scores on ProSem for Machine Translation (MT), cascade and direct Speech-to-Text Translation systems trained from scratch, as well random selection accuracy.}\label{tab:cont}
\end{table}

Table \ref{tab:cont} shows the contrastive evaluation overall accuracies for non-Whisper translation systems on the ProSem test set. The cascade model was not able to perform better than random, achieving a similar score but a higher score to the base gold MT. The base direct S2TT system could not outperform the cascade model, as its performance was weak overall as previously shown. In contrast, the ASR-initialised direct S2TT system outperformed the other systems, achieving an accuracy increase of 3.4\% over the cascade system. Although the overall accuracy remains modest, this observation lends credence to the hypothesis that direct S2TT systems effectively capture prosodic cues to disambiguate syntactically complex utterances.

\end{document}